\documentclass{article} % For LaTeX2e
\usepackage{iclr2027_conference,times}

\usepackage{amsmath,amsfonts,bm}

\def\eqref#1{equation~\ref{#1}}
\def\1{\bm{1}}

\DeclareMathAlphabet{\mathsfit}{\encodingdefault}{\sfdefault}{m}{sl}
\SetMathAlphabet{\mathsfit}{bold}{\encodingdefault}{\sfdefault}{bx}{n}

\DeclareMathOperator*{\argmax}{arg\,max}

\usepackage{booktabs}
\usepackage{graphicx}
\usepackage{hyperref}
\usepackage{url}
\usepackage{amssymb}
\usepackage{multirow}
\usepackage{tabularx}
\usepackage{adjustbox}
\usepackage{wrapfig}
\usepackage[capitalize,noabbrev]{cleveref}
\usepackage{enumitem}
\usepackage{tikz-cd}
\usepackage{graphicx}
\usepackage{pifont}
\usepackage{xcolor}

\title{Write Once, Query Many: Evolving Compact Reusable Memory for Long Videos}

\author{Kerui Chen$^{1,2}$, Jinglu Wang$^{2}$, Yun Wang$^{2}$, Xinjiang Wang$^{2}$, Yan Lu$^{2}$ \\
$^{1}$College of Artificial Intelligence, Zhejiang University. \\ 
$^{2}$Microsoft Research Asia \\
}

\newcommand{\ours}{Video-EMem}
\newcommand{\op}[1]{\operatorname{\mathsf{#1}}}
\newcommand{\cmark}{\ding{51}}
\newcommand{\xmark}{\ding{55}}
\newcommand{\gain}[1]{_{\textcolor{red}{+#1}}}

\begin{document}

\maketitle

\footnotetext[1]{This work was done when K. Chen was an intern at Microsoft Research Asia.}

\begin{abstract}
Long videos often exceed the context capacity of multimodal large language models (MLLMs), while agentic search repeats costly video processing and grounding for each query. Reusable memory can amortize this cost, but must preserve information for questions unknown during construction. We propose \ours{}, an evolvable video memory harness built on time-anchored perceptual observations. This representation combines compact free-form text with sparse images to preserve semantic content and complementary visual details across temporal scales, supporting temporal navigation without a fixed semantic schema. A Writer encodes each video once by consolidating overlapping multiscale observations, and a Reader retrieves and organizes evidence solely from the resulting memory. With the underlying models frozen, we evolve the operators within their writing and reading workflows using distinct feedback: query-independent source fidelity and coverage for the Writer, and QA failures in evidence localization and utilization for the Reader.
Experiments support this separation for generalization beyond development queries, with our evolution strategy improving average accuracy by \textbf{7.2\%} over the unevolved harness.
While reducing storage by about \textbf{99.86\%} on average relative to raw videos, \ours{} improves Qwen3-VL-30B by an average of {2.6\%} across three long-video QA benchmarks over direct input of 2,048 frames. Using the same o3 answer model as the state-of-the-art agentic method, it improves average accuracy by {2.5\%} with \boldsymbol{$19\times$} lower per-query latency.
\end{abstract}
\section{Introduction}
\label{sec:intro}

% Long video 

\begin{figure}
    \centering
    \includegraphics[width=1.0\linewidth]{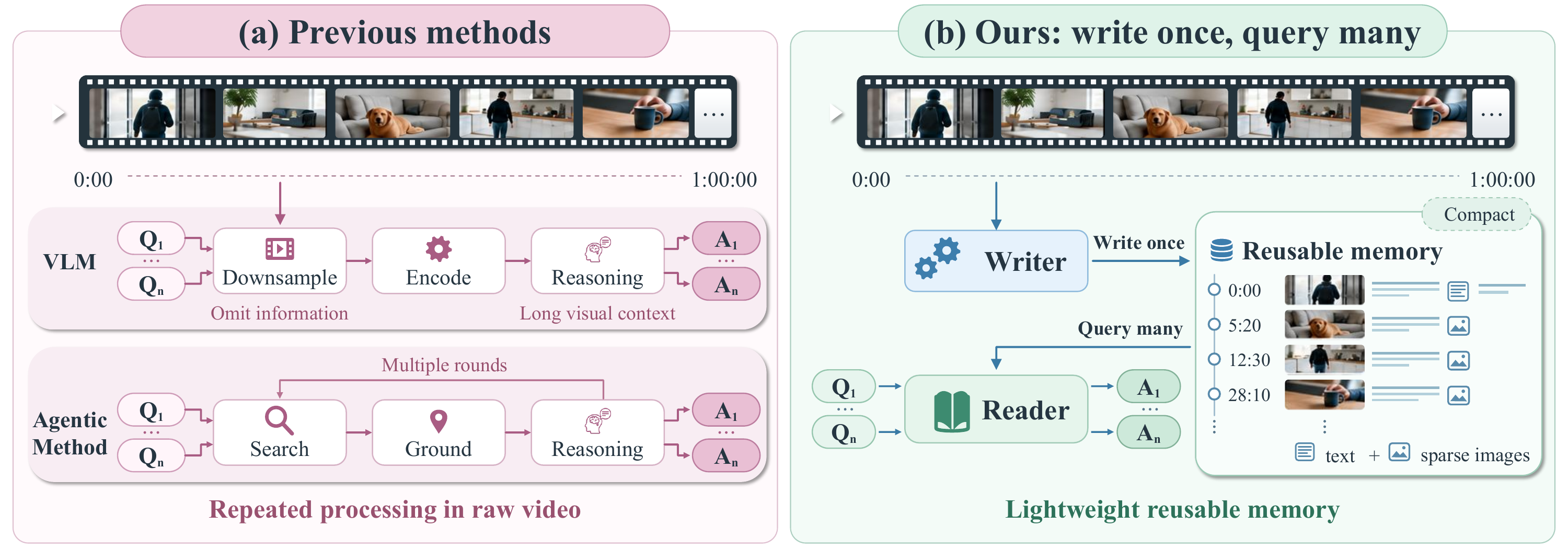}
    \vspace{-6mm}
    \caption{\textbf{The need for compact, reusable video memory.} (a) Answering multiple questions about the same video incurs repeated processing costs: VLMs repeatedly process long visual contexts and also often omit details due to context limits, while agentic methods revisit the raw video through multiple retrieval and reasoning rounds. (b) To reduce query and storage costs, we adopt a \emph{write-once, query-many} paradigm, encoding the video once into light-weight memory that preserves information for unknown future questions and supports efficient access across queries.}
    \label{fig:intro}
    \vspace{-6mm}
\end{figure}

Long videos often exceed the practical context capacity of multimodal large language models (MLLMs). Search-based and agentic methods address this limitation by iteratively locating relevant segments~\citep{ye2025re,fan2024videoagent,zhang2026deep, chen2026beyond}, but may repeat costly video processing and grounding, as shown in \cref{fig:intro} (a). Inspired by external memory in text-based agents~\citep{park2023generative,packer2023memgpt,wang2023augmenting}, an appealing alternative is to encode the video once into a compact representation reusable across queries. This raises a key question: \emph{how can we construct memory under a limited storage budget that preserves video information for unknown future questions and enables efficient evidence retrieval and organization?}

Existing video representations offer different trade-offs. Retaining all visual tokens is costly, while subsampling can discard details and disrupt temporal continuity~\citep{song2024moviechat,he2024ma,zhang2024flashvstream,yang2025streammem,wang2025internvideo2,qian2024videostreaming, chen2026scaling}. Textual memories are compact, but verbalization can omit fine-grained visual evidence~\citep{kahatapitiya2025language,ma2025drvideo}. Predefined schemas or relations for multimodal memory~\citep{fan2024videoagent,yin2026videoarm,yeo2026worldmm} may constrain what is retained, and independently processing temporal segments \citep{zhang2024simple, pang2025mr} can overlook broader context.  We therefore represent a video using a lightweight memory composed of \emph{time-anchored perceptual observations}. Rather than exhaustively storing all frames or intermediate observations, it selectively retains semantic and perceptual evidence across multiple temporal scales through free-form text and sparse images, without imposing a predefined semantic schema.

To construct and access such memory, we introduce \ours{}, an evolvable harness with frozen underlying MLLMs. Development queries may cover only a subset of video content, so query-guided optimization~\citep{xu2026long,wang2025videotree,diko2025rewind} risks overlooking evidence needed for unseen questions. Our harness therefore separates query-independent writing from query-specific reading. Although prior work adopts this separation~\citep{qian2024videostreaming,kahatapitiya2025language}, improving the two procedures under distinct objectives remains less explored. Our \emph{Writer} processes each video once, consolidating observations from overlapping, multiscale windows bottom-up. It compacts text within the token budget, calibrates source-supported information lost in compression, and supplements images while limiting redundancy. Our \emph{Reader} operates solely on the resulting memory: it retrieves text, accesses images for unresolved details, and organizes the evidence for a frozen answer VLM in a single retrieval-and-answer pass.

We leverage LLM-driven agent optimization~\citep{xu2026mem,zhang2026memskill,ye2026coevo,khattab2024dspy,opsahl2024optimizing,zhang2025aflow,agrawal2026gepa} to evolve the harness with separate feedback for each agent. Optimizing both agents solely for Question-Answering (QA) accuracy risks biasing memory toward development questions. The Writer therefore evolves using only source-video evidence, with feedback that rewards coverage and penalizes redundancy in textual and visual observations. With the resulting memory fixed, the Reader evolves from QA failures attributed to evidence localization or utilization. In both stages, an LLM makes targeted updates to the responsible operator.

We evaluate \ours{} on hour-long video QA benchmarks with a memory budget of at most 20K text tokens and 160 images per video, reducing storage by about $99.86\%$ on average relative to raw videos. For a one-hour video, this corresponds to $\sim 5.6$ tokens and $\sim 0.04$ images per second. Compared with direct video input, \ours{} improves Qwen3-VL-30B by 1.2\%, 4.8\%, and 1.9\% on LVBench~\citep{wang2025lvbench}, Video-MME Long~\citep{video-mme}, and LongVideoBench Long~\citep{wu2024longvideobench}, while the baseline using much more context (2048 images). The Writer evolves without QA supervision, while the Reader uses QA feedback only from a small subset of LVBench; both transfer to the other benchmarks without further adaptation. Our evolution strategy improves average accuracy by $7.2\%$ over the unevolved harness. In addition, compared with the state-of-the-art agentic method DVD~\citep{zhang2026deep}, \ours{} improves average accuracy by 2.5 percentage points (72.3\% vs.\ 69.8\%) with $\sim 19\times$ lower per-query latency (9.69\,s vs.\ 180\,s), using the same OpenAI o3 answer model. Our contributions are as follows:

\begin{itemize}[nosep, leftmargin=*]
    \item A compact, reusable video memory representation based on time-anchored perceptual observations that preserves source information independently of queries and supports efficient evidence retrieval and organization, and reduces storage by $\sim99.86\%$ relative to raw hour-long videos.
    \item An evolvable memory harness with separate Writer and Reader agents, optimized independently for source fidelity and QA performance to support generalization to unseen queries, improving average accuracy by 7.2\% over the unevolved baseline.
    \item Experiments demonstrating our improved long-video QA performance and query efficiency ($19 \times$ faster than agentic baseline).
\end{itemize}
\section{Related Work}
\noindent\textbf{Structured Video and Multimodal Memory.}
Memory has become a central mechanism for long-video understanding, enabling models to retain long-range evidence beyond their limited context windows and substantially improving performance. Early methods preserve video history through feature banks, recurrent states, or compressed visual tokens~\citep{wu2022memvit,song2024moviechat,he2024ma,zhang2024flashvstream,yang2025streammem}. Subsequent work externalizes video content into language repositories or searchable documents, making long-range evidence easier to retrieve and reason over~\citep{qian2024videostreaming,kahatapitiya2025language,ma2025drvideo}. More recent systems further organize text, images, objects, and events into structured multimodal memories that agents can query or update across temporal scales~\citep{fan2024videoagent,diko2025rewind,yin2026videoarm,yeo2026worldmm}. Despite this progress, memory construction typically follows a fixed architecture with predefined writing, compression, or update rules. Moreover, many repository- and agent-based memories accumulate observations with video duration or reasoning trajectories, while bounded approaches often focus on compressing latent or KV states rather than preserving broadly useful semantic and perceptual information. In contrast, \ours{} evolves its Writer from source-preservation feedback instead of manually specifying a fixed construction procedure. Under explicit text-token and image-count budgets, the Writer learns to preserve as much source information as possible, producing a lightweight, bounded, and reusable multimodal memory for future queries.

\noindent\textbf{Agentic Long-Video Understanding.}
Recent long-video understanding systems increasingly employ agentic workflows to locate and organize evidence beyond the context capacity of multimodal language models. Search-based methods progressively narrow the relevant temporal or spatial region: T*~\citep{ye2025re} performs temporal-spatial zooming, while DrVideo~\citep{ma2025drvideo} iteratively retrieves and augments a textual video document. VideoAgent~\citep{fan2024videoagent} equips a language-model agent with localization, memory-querying, and visual-perception tools, and DVD~\citep{zhang2026deep} uses agentic reasoning to repeatedly search long videos for question-relevant evidence. Related methods construct query-adaptive representations or retrieve relevant video segments, including R-VLM, VideoTree, and ReWind~\citep{xu2026long,wang2025videotree,diko2025rewind}.
Although these methods achieve superior performance, they largely overlook the substantial redundant computation involved. Instead of invoking the agent over the raw video for every question, \ours{} processes the video once into a lightweight, reusable memory, after which a Reader performs query-conditioned evidence access solely over the frozen memory. This write-once, query-many design substantially improves efficiency.

\noindent\textbf{LM-Based Memory and Workflow Evolution.}
Fixed memory operations and hand-designed LM workflows limit how agents improve from experience. Memory systems such as A-MEM~\citep{xu2026mem} reorganize stored content and links, while MemSkill~\citep{zhang2026memskill} learns and refines reusable memory operations. CoEvo-Mem~\citep{ye2026coevo} further alternates updates to a retrieval policy and memory bank using task outcomes. A related line optimizes prompts, modular LM programs, or code-represented workflows from execution feedback, including DSPy, MIPRO, AFlow, and GEPA~\citep{khattab2024dspy,opsahl2024optimizing,zhang2025aflow,agrawal2026gepa}. In this paper, to address the dual challenges of constructing a broadly informative video memory before future questions are known and efficiently accessing relevant evidence from it for each query, we adopt a two-stage evolution process. First, we evolve a query-independent Writer using source-preservation feedback and materialize the resulting video memories. We then freeze these memories and evolve a query-conditioned Reader from attributed downstream QA failures.

\section{Method}
\label{sec:method}

\subsection{Problem Setting}
\label{sec:problem-setting}
\begin{figure}
    \centering
    \includegraphics[width=0.95\linewidth]{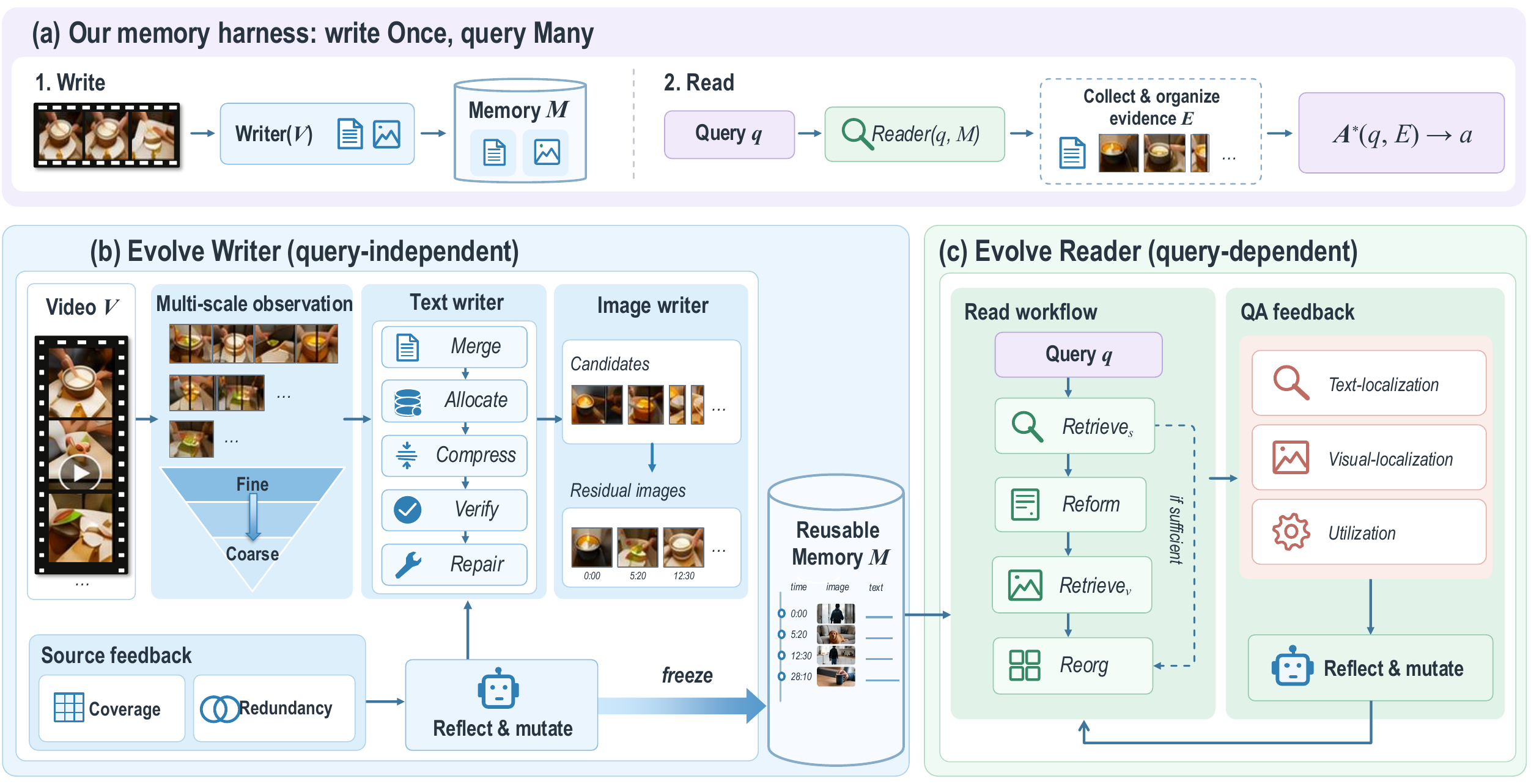}
    \vspace{-3mm}
    \caption{\textbf{Method overview.} (a) We present \ours{}, an evolvable memory harness with a Writer that constructs compact, query-independent video memory and a Reader that retrieves and organizes evidence for question answering. Our memory consists of time-anchored observations combining multiscale textual descriptions with sparse complementary images. We first evolve the Writer (b) using source-preservation feedback without queries, then evolve the Reader (c) from QA failures attributed to evidence localization and utilization.}
    \label{fig:method}
\end{figure}

Given a long video $V$ and a set of queries $Q$, conventional video understanding methods predict an answer $a=A(q,V)$ for each query $q\in Q$, incurring substantial computational costs from repeated visual encoding and long-context agentic reasoning. We instead study a \emph{write-once, query-many} setting: the video is processed once into a compact and lightweight reusable memory $M$, from which all subsequent queries are answered without revisiting the raw video. To realize this setting, we introduce \ours{}, consisting of
two agents: a \emph{Writer} constructs $M$ from $V$, while a \emph{Reader} accesses the frozen memory for each downstream query, as illustrated in \cref{fig:method} (a). The key insight behind it is that memory construction and memory access operate under different conditions and therefore need the different optimization goals. 

\noindent\textbf{Writer objective.}
The Writer aims to construct a query-independent memory $M$ under a budget $B$ before observing any downstream queries. Because future information needs are unknown, its objective is to preserve a broad range of source information to support diverse queries:
\begin{equation}
\max_{\operatorname{Writer},\,M}
\ \operatorname{Info}(V,M)
\qquad
\text{subject to}
\qquad
M=\operatorname{Writer}(V;B),\quad |M|\leq B,
\label{eq:writer-objective}
\end{equation}
where $|M|$ denotes memory usage under the prescribed accounting scheme, and $\operatorname{Info}(V,M)$ quantifies the source information retained in $M$. The construction and evolution procedures introduced below realize this objective through source-preservation diagnostics. Once constructed, $M$ is fixed and reused across all downstream queries.

\noindent\textbf{Reader objective.}
For each query $q$, the Reader selects and organizes relevant evidence from the frozen memory $M$ for an answer model $A^\star$. Unlike the Writer, which constructs memory independently of downstream questions, the Reader is optimized directly from question-answer supervision:
\begin{equation}
\min_{\operatorname{Reader}}\ \mathbb{E}_{(M,q,\hat a)\sim\mathcal D_{\mathrm{QA}}}
\left[
\mathcal L_{\mathrm{QA}}\!\left(
A^\star\!\left(q,\operatorname{Reader}(q,M)\right),\hat a
\right)
\right],
\label{eq:reader-objective}
\end{equation}
where $\mathcal D_{\mathrm{QA}}$ denotes the training set of memory-query-answer tuples and $\mathcal L_{\mathrm{QA}}$ measures the prediction error with respect to the ground-truth answer $\hat a$. During Reader optimization, both the memory $M$ and the answer model $A^\star$ remain fixed. 

\subsection{Reusable Video Memory Representation}
Reusable video memory must balance information fidelity, storage cost, and evidence accessibility. Dense visual tokens are costly to retain for long videos, while compact textual descriptions may omit visual details that are difficult to verbalize.

We therefore adopt a \emph{hybrid multimodal memory representation}: a temporally ordered sequence of time-anchored observations, each comprising a compact textual record and, where needed, sparse images that preserve complementary visual observations within a temporal interval. This representation provides a shared interface for the Writer to encode video content and the Reader to retrieve relevant evidence.

\noindent\textbf{Time-anchored multimodal observations.}
We represent memory $M$ as a sequence of $N$ observations ordered by their interval start times:
\vspace{-3mm}
\begin{equation}
M=[o_1,o_2,\ldots,o_N], o_i=(\Delta_i,s_i,\mathcal I_i), \quad
\text{subject to}\quad
\sum_{i=1}^{N}|s_i|_{\mathrm{tok}}\leq B_s,\quad
\sum_{i=1}^{N}|\mathcal I_i|\leq B_v,
\label{eq:memory-representation}
\vspace{-2mm}
\end{equation}
Here, $\Delta_i=[t_i^{\mathrm{start}},t_i^{\mathrm{end}}]$ denotes the video interval, $s_i$ its textual record, and $\mathcal{I}_i$ a sparse or possibly empty set of timestamped images from that interval. The memory budget $B=(B_s,B_v)$ separately limits text tokens and image counts to account for their different encoding costs, with $|s_i|_{\mathrm{tok}}$ denoting the token count of $s_i$.
 
The textual record $s_i$ forms the primary memory scaffold, integrating information from multiple temporal scales: fine-grained observations capture local actions and state changes, while coarser observations capture event progression and longer-range context. The visual component $\mathcal{I}_i$ supplements this record with sparse evidence of appearance, spatial layout, and subtle or transient details that are difficult to verbalize. Image timestamps align this evidence with the text, and no images are retained when the textual record is sufficient.

\subsection{Writer: Constructing Query-Independent Memory}
The Writer first constructs textual observations from a multi-scale source corpus, then supplements them with sparse images that provide complementary visual observations. The writing workflow and evolution are illustrated in \cref{fig:method} (b).

\subsubsection{Writing Workflow}
In the writing workflow, we first construct textual observations from multi-scale source corpus and then supplement the textual with very sparse images to provide visual observations.

\noindent\textbf{Multi-scale source corpus.}
To capture fine-grained details and long-range context, we construct a multi-scale caption corpus for the Writer. At each temporal scale $\ell \in \{1,\ldots,L\}$, ordered from coarsest to finest, we divide the video into intervals $\mathcal{T}^{\ell}$ and independently generate a caption $c_{\Delta^\ell}$ for each interval $\Delta^\ell \in \mathcal{T}^{\ell}$. These captions provide complementary descriptions at different temporal granularities and serve as the Writer's source corpus.

\noindent\textbf{Compact.}
The Writer consolidates textual observations from fine to coarse scales. For each interval $\Delta^\ell$, to produce the final observation $z_{\Delta^\ell}$, we merges the source caption $c_{\Delta^\ell}$ with the observations $z_{\Delta^{\ell+1}}$ from finer-scale intervals contained within it ($\Delta^{\ell+1}\subseteq\Delta^\ell$), allocates the text-token budget across the merged observations, and compresses each observation to its assigned capacity:
\begin{equation}
z'_{\Delta^\ell}
= \op{Merge}\!\left(
c_{\Delta^\ell},
\{z_{\Delta^{\ell+1}}\}_{\Delta^{\ell+1}\subseteq\Delta^\ell}
\right),
\quad
\mathbf{b}_{\ell}
= \op{Allocate}\!\left(\mathbf{z}'_{\ell};B_s\right),
\quad
z''_{\Delta^\ell}
= \op{Compress}\!\left(z'_{\Delta^\ell};b_{\Delta^\ell}\right).
\label{eq:compact-operators}
\end{equation}
Here, $\mathbf{z}'_{\ell}$ is the temporally ordered sequence of merged observations at scale $\ell$, and $\mathbf{b}_{\ell}$ contains their allocated token budgets, satisfying $\sum_{\Delta^\ell \in \mathcal{T}^{\ell}} b_{\Delta^\ell} \leq B_s$. At the finest scale, merging uses only the source caption. Budget allocation assigns less capacity to redundant content and more to distinctive information.

\noindent\textbf{Calibrate.}
Compression may omit or weaken source information. Before the representation is propagated to the next scale, the Writer verifies it against the source sentences before merge:
\begin{equation}
r_{\Delta^\ell}
=
\op{Verify}(z''_{\Delta^\ell}, c_{\Delta^\ell},\{z_{\Delta^{\ell+1}}\}_{\Delta^{\ell+1}\subseteq\Delta^\ell}),
\qquad
z_{\Delta^\ell}
=
\op{Repair}(z''_{\Delta^\ell},r_{\Delta^\ell};b_{\Delta^\ell}),
\label{eq:calibrate-operators}
\end{equation}
where $r_{\Delta^{\ell+1}}$ records source-supported information that is missing or weakened after compression. $\op{Repair}$ selectively restores omissions within the assigned token budget.
The calibrated observation $z_{\Delta^\ell}$ is passed to the next coarser scale when $\ell>1$. At $\ell=1$, the outputs form the final textual observations $\{s_i\}_{i=1}^{N}$, where $s_i=z_{\Delta_i^1}$.

\noindent\textbf{Visual supplement.}
With the textual memory consolidated, the Writer constructs visual observations by selecting images that complement the text while limiting visual redundancy. For each textual observation $s_i$, we initialize the selected image set $\mathcal{I}_i=\varnothing$ and greedily update it as
\begin{equation}
\mathcal{I}_i \leftarrow \mathcal{I}_i \cup \{I^\star\},
\quad
I^\star
=
\arg\max_{I}
\left[
\rho_{\mathrm{MLLM}}(I,s_i)
-
\lambda\rho_{\mathrm{DINO}}(I,\mathcal{I}_i)
\right].
\label{eq:residual-image-selection}
\end{equation}
The maximization ranges over unselected candidate images within the interval of $s_i$, with scores computed before each update. Here, $\rho_{\mathrm{MLLM}}$ estimates the information an image adds beyond the text with an MLLM, and $\rho_{\mathrm{DINO}}$ measures its maximum DINO-feature cosine similarity to selected images. The parameter $\lambda$ controls the redundancy penalty. Selection continues until the allocated image budget is reached. Each selected image retains its timestamp for localization.

\subsubsection{Writer Evolution}
We evolve the Writer using source-preservation feedback without downstream QA supervision. Following the memory construction order, we first optimize and freeze text construction, then optimize visual selection to complement the fixed text. Both stages use the mutation strategy summarized below and detailed in~\cref{app:guarded-mutation}.

We guide text construction with two diagnostics: coverage and redundancy.
(1) The \emph{coverage diagnostic} identifies information lost during construction by comparing the final textual records with the source captions and decomposing omissions into atomic facts. For each video, we retain the construction trajectory with the highest proportion of missing facts, together with those facts, to guide updates to the responsible operators.
(2) The \emph{redundancy diagnostic} identifies equivalent, subsumed, or partially overlapping content across the final records. We check temporal locations and source-caption provenance to distinguish redundant descriptions from distinct recurring events. We then retain the groups whose consolidation would save the most tokens, together with their construction trajectories, to guide updates that reduce repetition while preserving distinct evidence.

With the text-writing procedure fixed, we optimize the Writer's image selection to complement the text with visual observations while reducing redundancy. An MLLM classifies each retained image as \emph{complementary}, \emph{text-overlapping}, or \emph{uninformative} relative to its textual observation, while DINO-feature similarity measures redundancy among images. We aggregate these diagnostics to identify the dominant selection failure and guide targeted updates to residual-value scoring, redundancy penalization, or the retention rule. Each update is evaluated using a complementarity score.

\subsection{Reader: Accessing Constructed Memory}
After the Writer evolution, we construct and freeze video memories, then evolve the Reader using memory--question--answer tuples in a training set. The reading workflow and evolution are illustrated in \cref{fig:method} (c).

\subsubsection{Reading Workflow}
Given a query $q$, the Reader retrieves text, reformulates the query to retrieve images for unresolved details, and reorganizes the evidence into context:
\begin{equation}
\resizebox{0.94\linewidth}{!}{$\displaystyle
\mathcal{E}_s=\op{Retrieve}_s(q,\{s_i\})
\to
q'=\op{Reform}(q,\mathcal{E}_s)
\to
\mathcal{E}_v=\op{Retrieve}_v(q',\{\mathcal{I}_i\})
\to
E=\op{Reorg}(q,\mathcal{E}_s,\mathcal{E}_v).
$}
\end{equation}
If the text is sufficient, $\op{Reform}$ and $\op{Retrieve}_v$ are skipped, with $\mathcal{E}_v=\varnothing$. The frozen answer model generates the answer as $a=A^\star(q,E)$.

\subsubsection{Reader Evolution}
We evolve the Reader from failed QA attempts. For each incorrectly answered pair $(q,\hat{a})$, we use the reference temporal interval $\hat{\Delta}$ and the frozen memory to attribute the failure to the responsible operator: (1)  \emph{Text-localization failure}: relevant text associated with $\hat{\Delta}$ exists in memory but is missing from $\mathcal{E}_s$, implicating $\op{Retrieve}_s$. (2)  \emph{Visual-localization failure}: an image supporting $\hat{a}$ exists in memory but absent from $\mathcal{E}_v$, implicating $\op{Reform}$ or $\op{Retrieve}_v$. (3) \emph{Evidence-utilization failure}: retrieved evidence is obscured by redundant, irrelevant, or poorly ordered context, implicating $\op{Reorg}$.

Information absent from memory is not counted as a Reader localization failure. An LLM diagnoses each attributed failure using the operator's execution trajectory, question, prediction, and reference evidence. These diagnoses are aggregated across questions to identify recurring issues and guide revisions in the next round.

\noindent\textbf{Unified workflow mutation.}
The Writer and Reader share a mutation procedure. Each round, a coding agent uses diagnostic feedback to edit the responsible operator, then evaluates the revision with the stage-specific score while keeping other components and constraints fixed. A sharp score drop triggers rollback to the best previous version. After two rounds of little score change, the agent proposes an exploratory edit beyond the latest diagnosis to help escape a local optimum. Evolution stops at the maximum number of rounds; details appear in Appendix~\cref{app:guarded-mutation}.
% \clearpage
\section{Experiment}

\subsection{Experimental Setup}
\noindent\textbf{Benchmarks and metric.}
We report accuracy (\%) on three long-video QA benchmarks. LVBench~\citep{wang2025lvbench} contains 103 videos averaging 68 minutes. Video-MME~\citep{video-mme} contains 900 videos and 2,700 questions across diverse domains and durations; we evaluate its 30-60 minutes subset without subtitles, denoted Video-MME-L. LongVideoBench~\citep{wu2024longvideobench} contains 3,763 videos and 6,678 human-annotated questions; we evaluate the long-video subset of its validation set, denoted LongVideoBench-L.

\noindent\textbf{Evolution and evaluation splits.}
We reserve 30 LVBench videos for development, maintaining video-level separation from evaluation. Of these, 20 provide query-independent source feedback for Writer evolution and QA feedback for Reader evolution. We use all the 30 development videos to select the final programs. The other 73 LVBench videos are held out for evaluation. All LVBench results in this paper are evaluated on the same 73 held-out videos. We additionally evaluate on Video-MME-L and LongVideoBench-L without further adaptation to assess cross-benchmark generalization.

\textbf{Baselines.}
We compare \textsc{\ours} with two complementary groups of strong baselines. The first group comprises three state-of-the-art VLMs, Gemini-2.0-Flash, OpenAI o3, and Qwen3-VL-30B, representing direct long-context video understanding. The second comprises three state-of-the-art agentic methods, PACE~\citep{xu2026finding}, MR. Video~\citep{pang2025mr}, and DVD~\citep{zhang2026deep}, which iteratively access the source video to gather question-relevant evidence. Comparing with the agentic methods, we use the same answer model as each corresponding baseline, thereby isolating differences in memory construction and video access.

\textbf{Models and budgets.}
GPT-4.1 serves as the frozen captioner and writer executor, GPT-5.3-codex as the reflector for both evolution stages, and Qwen3-VL-30B-A3B-Instruct as the default frozen answer model. Unless otherwise stated, our memory contains at most 20K text tokens and 160 images stored at $448\times448$ resolution. Empirically, the resulting memory reduces storage by $\approx 99.86\%$ on average, occupying only $0.14\%$ of the raw-video storage. All model weights and selected programs remain fixed at test time.

\begin{table}[t]
    \caption{\textbf{Comparison on long-video benchmarks}, reporting accuracy (\%) and query time (s). ``Avg'' is the mean accuracy across the three benchmarks. ``Reusable Limits'' specifies memory storage budgets. VLMs receive the full stored representation as input, while our method supplies only a query-relevant memory subset to the answer model.}
    \centering
    \renewcommand\arraystretch{1.15}
    \begin{adjustbox}{width=0.95\linewidth}
    \begin{tabular}{l cccccc}
        \toprule
        Methods & Reusable Limits & LVBench & Video-MME-L & LongVideoBench-L & Avg. & Time (s) \\
        \midrule
        \multicolumn{7}{l}{\textcolor{gray}{\textit{Frontier VLMs}}} \\
        Gemini-2.0-Flash & 256 images & 48.4 & 63.0 & 45.7 & 52.4 & -- \\
        OpenAI o3 & 256 images & 56.8 & 64.7 & 60.6 & 60.7 & 12.9 \\
        Qwen3-VL-30B & 2048 images & 61.9 & 61.6 & 59.2 & 60.9 & 48.7 \\
        \midrule
        \multicolumn{7}{l}{\textcolor{gray}{\textit{Agentic Methods}}} \\
        PACE (Qwen3-VL-30B) & -- & 61.4 & 58.9 & 55.1 & 58.5 & $\sim$3000 \\
        MR. Video (gpt-4o) & -- & 60.2 & 61.8 & 61.6 & 61.2 & $\sim$120 \\
        DVD (o3) & -- & 73.5 & 67.3 & 68.6 & 69.8 & $\sim$180 \\
        \midrule
        ours-Text (Qwen3-VL-30B) & $\leq$20K text & 61.3 & 64.9 & 59.7 & 62.0 & 3.76 \\
        ours (Qwen3-VL-30B) & $\leq$20K text + 160 images & 63.1 & 66.4 & 61.1 & 63.5 & 5.51 \\
        ours (gpt-4o) & $\leq$20K text + 160 images & 63.4 & 65.5 & 61.8 & 63.6 & 4.42 \\
        ours (o3) & $\leq$20K text + 160 images & 74.4 & 72.8 & 69.7 & 72.3 & 9.69 \\
        \bottomrule
    \end{tabular}
    \end{adjustbox}
    \vspace{-6mm}
    \label{tab:main}
\end{table}

\subsection{Main Results}
\cref{tab:main} compares \ours{} with frontier VLMs and agentic methods for long-video understanding. Following the conventional evaluation setting, VLM baselines receive 256 and 2048 sampled frames. Agentic baselines access the original video through multiple rounds of retrieval and inference, rather than relying on a compressed reusable memory. 

\noindent\textbf{Superior performance.}
With at most 20K text tokens and 160 images per video, \ours{} achieves the highest accuracy among the compared methods on all three benchmarks: 74.4\% on LVBench, 72.8\% on Video-MME, and 69.7\% on LongVideoBench using OpenAI o3 as the answer model. These results demonstrate the effectiveness of our memory representation across diverse long-video QA benchmarks. Moreover, \ours{} consistently outperforms prior agentic methods with the same answer model, supporting the contribution of memory construction and reuse beyond improvements in model capability.

\begin{table}[t]
    \centering
    \footnotesize
    \caption{\textbf{Effect of evolution strategies.}
    \cmark/\xmark~indicate whether evolution is performed.
    }
    \label{tab:two_timescale}
    \resizebox{0.7\linewidth}{!}{%
    \begin{tabular}{cccccc}
        \toprule
        Writer & Reader & LVBench & Video-MME-L & LongVideoBench-L & Avg. \\
        \midrule
        \xmark & \xmark
        & $54.2$ & $60.3$ & $54.4$ & $56.3$ \\
        \cmark & \xmark
        & $60.3\gain{6.1}$ & $64.7\gain{4.4}$ & $59.2\gain{4.8}$
        & $61.4\gain{5.1}$ \\
        \xmark & \cmark
        & $56.4\gain{2.2}$ & $61.5\gain{1.2}$ & $55.6\gain{1.2}$
        & $57.8\gain{1.5}$ \\
        \multicolumn{2}{c}{Joint QA Evolution}
        & $57.5\gain{3.3}$ & $62.5\gain{2.2}$ & $55.7\gain{1.3}$
        & $58.6\gain{2.3}$ \\
        \cmark & \cmark
        & $\mathbf{63.1}\gain{8.9}$
        & $\mathbf{66.4}\gain{6.1}$
        & $\mathbf{61.1}\gain{6.7}$
        & $\mathbf{63.5}\gain{7.2}$ \\
        \bottomrule
    \end{tabular}%
    }
    \vspace{-6mm}
\end{table}

\noindent\textbf{Query efficiency.}
\ours{} combines strong accuracy with low per-query latency by accessing preconstructed memory and invoking the answer model only once. The Qwen3-VL-30B variant takes 5.51\,s per query, compared with approximately 3,000\,s for PACE. With OpenAI o3, \ours{} takes 9.69\,s, compared with approximately 180\,s for DVD. These results demonstrate that memory reuse supports efficient querying while preserving information needed for accurate answers.

\begin{table*}[t]
\centering

\vspace{0pt}
\begin{minipage}[t]{0.48\linewidth}
\centering
\caption{\textbf{Effect of the writer operators for text.}}
\label{tab:writer_ablation}
\begin{adjustbox}{width=\linewidth}
\footnotesize
\begin{tabular}{lccc}
\toprule
Writer & LVBench & Video-MME-L & LongVideoBench-L\\
\midrule
w/o $\op{Merge}$  & 57.8 & 62.2 & 57.5 \\
w/o $\op{Calibrate}$   & 59.3 & 63.6 & 57.8\\
w/o $\op{Allocate}$   & 60.5 & 63.8 & 59.1 \\
Full          & 61.3 & 64.9 & 59.7 \\
\bottomrule
\end{tabular}
\end{adjustbox}
\end{minipage}
\hfill
\begin{minipage}[t]{0.48\linewidth}
\centering
\vspace{0pt}
\caption{\textbf{Effect of visual supplement (\scriptsize {$\op{VS}$).}}}
\label{tab:visual_selection}
\begin{adjustbox}{width=\linewidth}
\footnotesize
\begin{tabular}{lcccc}
\toprule
$\op{VS}$ & Frames & LVBench & Video-MME-L & LongVideoBench-L\\
\midrule
- & - & 61.3 & 64.9 & 59.7 \\
\midrule
Uniform & 80  & 61.5 & 65.3 & 59.8\\
Residual & 80  & 62.1 & 65.8 & 60.6 \\
\midrule
Uniform & 160 & 61.7 & 65.6 & 60.3 \\
Residual & 160 & 63.1 & 66.4 & 61.1\\
\bottomrule
\end{tabular} 
\end{adjustbox}
\end{minipage}

\vspace{-4.2mm}
\end{table*}

\subsection{Ablation Study}
\noindent\textbf{Effect of Writer and Reader evolution.}
\cref{tab:two_timescale} shows that Writer and Reader 
evolution provide 
\begin{wrapfigure}{r}{0.33\linewidth}
    \centering
    \vspace{-4mm}
    \includegraphics[width=\linewidth]{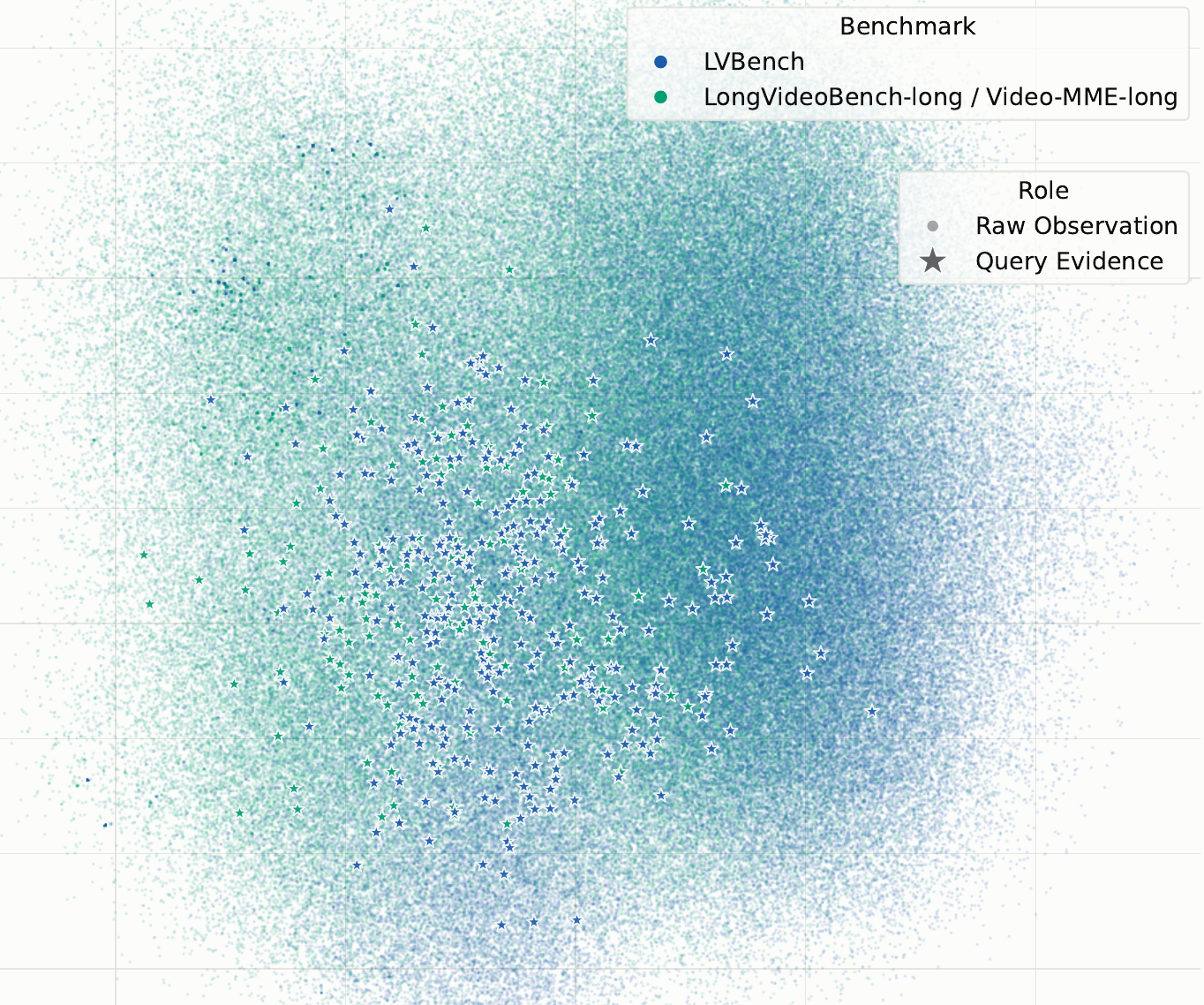}
    \caption{\textbf{Query-relevant evidence (white stars) cover only a limited portion of source information (blue and green dots).}}
    \label{fig:embedding}
    \vspace{-6mm}
\end{wrapfigure}
complementary gains. 
Evolving only the Writer improves accuracy by 6.1, 4.4, and 4.8 points on LVBench, Video-MME, and LongVideoBench, respectively. Reader evolution adds 2.2, 1.2, and 1.2 points with the fixed Writer, and larger gains of 2.8, 1.7, and 1.9 points with the evolved Writer. This suggests that better memory construction enables more effective query-specific access. Separately evolving both agents achieves the best results, outperforming joint QA evolution by 5.6, 3.9, and 5.4 points and supporting the use of distinct objectives for writing and reading.

\noindent\textbf{Cross-benchmark generalization.}
The programs evolved on LVBench transfer to Video-MME and LongVideoBench without adaptation, achieving 66.4\% and 61.1\% accuracy, respectively. These results improve over the unevolved pipeline by 6.1 and 6.7 points, suggesting that the learned memory construction and access strategies generalize beyond the development benchmark.

\cref{fig:embedding} qualitatively illustrates the relationship between source information and evidence relevant to development queries. We project source-caption embeddings from LVBench and other benchmarks, highlighting those associated with the development queries. The highlighted observations represent only part of the source content, motivating source-preservation feedback beyond the observed QA pairs. This projection is illustrative rather than a quantitative measure of information coverage.

\cref{tab:two_timescale} provides complementary empirical support: separate Writer and Reader evolution consistently outperforms joint QA evolution across all three benchmarks. These results support using distinct objectives for memory construction and access, although they do not by themselves establish that the joint strategy overfits development queries.

\noindent\textbf{Effect of Writer operators.}
We evaluate text-only memory to isolate the contribution of the Writer's textual operators in \cref{tab:writer_ablation}. $\op{Merge}$ preserves multi-scale source information during hierarchical consolidation, while $\op{Allocate}$ assigns capacity according to content importance rather than compressing observations uniformly. $\op{Calibrate}$ combines $\op{Verify}$ and $\op{Repair}$ to detect and restore information lost or distorted during compression. Together, these operators achieve the best balance of coverage, redundancy, and QA accuracy.

\begin{figure}[t]
    \centering
    \includegraphics[width=0.9\linewidth]{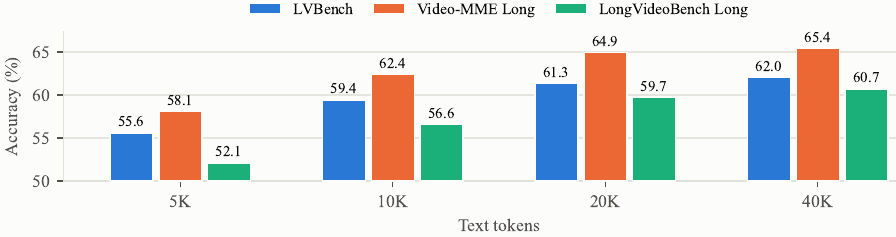}
    \vspace{-4mm}
    \caption{\textbf{Effect of text-memory budgets.} Increasing the budget from 5K to 20K tokens substantially improves accuracy, while expanding it to 40K yields only marginal gains. This saturation may partly reflect visual details omitted from source captions, motivating complementary images.}
    \label{fig:memory_budget}
    \vspace{-3.5mm}
\end{figure}

\noindent\textbf{Effect of text budget.}
\cref{fig:memory_budget} shows that increasing the text budget from 5K to 20K tokens improves mean accuracy across the three benchmarks by 6.7 points, while doubling it to 40K adds only 0.7 points. These diminishing gains may partly reflect limitations of the captioning model: additional text capacity cannot recover visual details absent from the source captions, motivating complementary visual observations. We therefore use 20K tokens by default to balance accuracy and memory cost, equivalent to approximately 5.6 tokens per second for a one-hour video.

\begin{figure}
    \centering
    \includegraphics[width=0.9\linewidth]{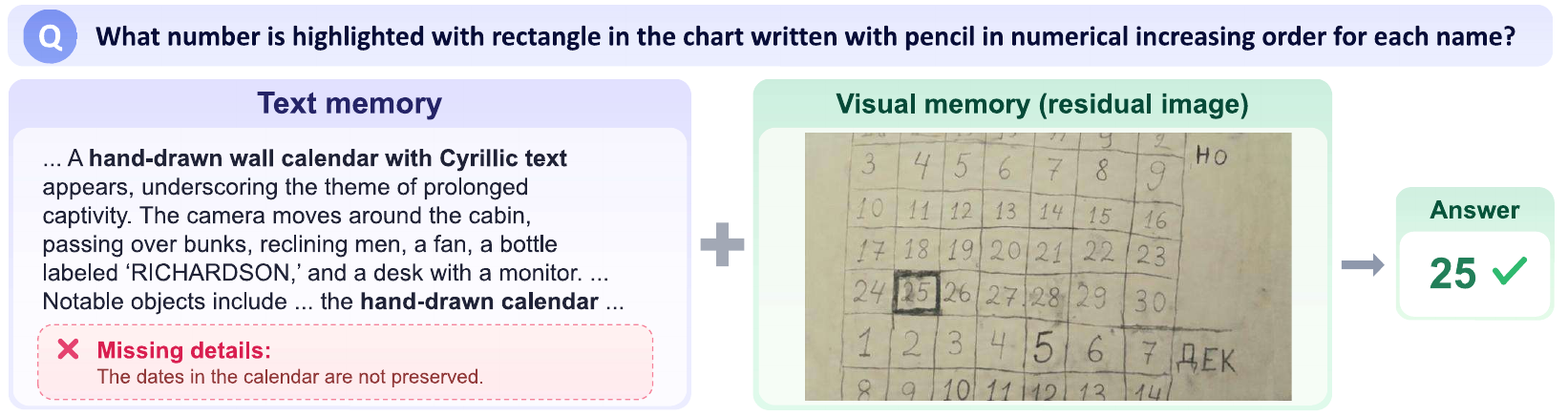}
    \vspace{-4mm}
    \caption{\textbf{Residual images preserve visual details omitted from textual memory.} In this example, the textual observations describe the cluttered scene but omit the marked number. The retained image preserves these details, enabling the correct answer.}
    \label{fig:visual_mem}
    \vspace{-5mm}
\end{figure}

\noindent\textbf{Effect of visual observations.}
\cref{tab:visual_selection} shows that residual selection outperforms uniform sampling at the same image budget. With 160 images, it achieves 63.1\% accuracy on LVBench versus 61.7\% for uniform sampling. Even 80 residual images achieve 62.1\%, outperforming 160 uniformly sampled images with half the storage. This demonstrates the benefit of selecting images that complement textual memory.
For a one-hour video, 160 images correspond to only about 0.04 frames per second, yet preserve useful visual details omitted from text. 
In ~\cref{fig:visual_mem}, the textual memory describes a hand-drawn calendar but omits its marked numbers; the retained image supplies these details and enables the correct answer. 
\section{Conclusion}
In this paper, we introduced \ours{}, an evolvable write-once, query-many memory harness for efficient long-video understanding. It encodes each video into compact, time-anchored perceptual observations that combine free-form text with sparse complementary images. With the underlying MLLMs frozen, \ours{} separately evolves a query-independent Writer using source-preservation feedback and a query-specific Reader using attributed QA failures. Experiments on three long-video QA benchmarks demonstrate improved accuracy, strong cross-benchmark generalization, and substantially lower per-query latency than direct-input and agentic baselines.

\subsection*{AI use statement}
In this work, generative AI models were used as components of the proposed method and experiments, including video captioning, memory construction, workflow evolution, and answer generation, as detailed in the method and experimental sections. These models were not used to design the research methodology or prepare the manuscript. We also used ChatGPT solely to improve the grammar, clarity, and readability of the manuscript. ChatGPT was not used to generate substantive scientific content, including research hypotheses, methodology, analyses, interpretations, claims, or conclusions. Other tasks listed as requiring disclosure under the ICLR policy were either not performed with generative AI or were not applicable to this work. We take full responsibility for the experimental design, analysis, and final content of the paper.

\bibliography{iclr2027_conference}
\bibliographystyle{iclr2027_conference}

\newpage
\appendix
\section*{Appendix}
\section{Additional Method Details}
\label{app:method-details}

\subsection{Detailed Visual-Supplement Procedure}
\label{vis:supple}

The visual-supplement stage is applied after the textual records $\{s_i\}_{i=1}^{N}$ have been finalized. Rather than summarizing the video with representative frames, it retains sparse visual evidence that complements the information already preserved in text. The procedure below expands the greedy selection rule in \cref{eq:residual-image-selection}, using the memory representation $o_i=(\Delta_i,s_i,\mathcal I_i)$ and the global image-count budget $B_v$ defined in \cref{eq:memory-representation}.

\paragraph{Candidates and selection scores.}
For each observation $o_i$, let
\begin{equation}
  \mathcal C_i
  =
  \left\{
  (t_{i,j},I_{i,j})
  \right\}_{j=1}^{m_i},
  \qquad t_{i,j}\in\Delta_i,
  \label{eq:appendix-image-candidates}
\end{equation}
denote the candidate timestamp--image pairs from interval $\Delta_i$. The score $\rho_{\mathrm{MLLM}}(I,s_i)$ estimates the residual information that image $I$ provides beyond textual record $s_i$. Consistent with the main text, we define visual redundancy as
\begin{equation}
  \rho_{\mathrm{DINO}}(I,\mathcal I_i)
  =
  \begin{cases}
  0,
  & \mathcal I_i=\varnothing,\\[1mm]
  \displaystyle
  \max_{(t',I')\in\mathcal I_i}
  \operatorname{sim}\!\left(
  f_{\mathrm{DINO}}(I),
  f_{\mathrm{DINO}}(I')
  \right),
  & \text{otherwise},
  \end{cases}
  \label{eq:appendix-visual-redundancy}
\end{equation}
where $f_{\mathrm{DINO}}$ is the frozen DINO encoder and $\operatorname{sim}(\cdot,\cdot)$ denotes cosine similarity in its feature space. At each step, the Writer selects
\begin{equation}
  (t^\star,I^\star)
  =
  \argmax_{(t,I)\in\mathcal C_i\setminus\mathcal I_i}
  \left[
  \rho_{\mathrm{MLLM}}(I,s_i)
  -
  \lambda\rho_{\mathrm{DINO}}(I,\mathcal I_i)
  \right],
  \label{eq:appendix-image-utility}
\end{equation}
and updates $\mathcal I_i\leftarrow\mathcal I_i\cup\{(t^\star,I^\star)\}$, subject to $\sum_i|\mathcal I_i|\leq B_v$. Here, $\lambda$ controls the penalty for visual redundancy.

\subsection{Writer-Evolution Diagnostics}
\label{app:writer-evolution-diagnostics}

Writer evolution follows the same order as memory construction. We first optimize text construction, materialize and freeze the resulting textual records $\{s_i\}_{i=1}^{N}$, and then optimize image selection conditional on those fixed records. The diagnostics below evaluate the output of a candidate Writer; they are not additional construction operators. In particular, diagnostic source coverage is distinct from the construction-time $\op{Verify}$ operation in \cref{eq:calibrate-operators}.

\subsubsection{Text-Writer Diagnostics}

For each final record $s_i=z_{\Delta_i^1}$, we retain the construction trajectory that produced it. A trajectory contains, for every contributing interval $\Delta^\ell\in\mathcal T^\ell$, its source caption $c_{\Delta^\ell}$, merged state $z'_{\Delta^\ell}$, assigned capacity $b_{\Delta^\ell}$, compressed state $z''_{\Delta^\ell}$, verification result $r_{\Delta^\ell}$, and calibrated state $z_{\Delta^\ell}$. At each scale, the associated allocation context $\mathbf z'_\ell$ and $\mathbf b_\ell$ is also retained. These traces connect a loss or redundancy observed in $s_i$ to the decisions of $\op{Merge}$, $\op{Allocate}$, $\op{Compress}$, $\op{Verify}$, and $\op{Repair}$.

\paragraph{Source coverage.}
Let $\mathcal S_i$ be the source-caption sentences that contribute to final record $s_i$, and let $\operatorname{Sent}(s_i)$ be the sentences retained in that record. We define the diagnostic coverage indicator
\begin{equation}
c(u,s_i)
=
\mathbf 1\!\left[
\exists m\in\operatorname{Sent}(s_i):
\operatorname{Match}(u,m)=1
\right],
\qquad u\in\mathcal S_i,
\label{eq:appendix-source-coverage}
\end{equation}
where $\operatorname{Match}$ is the semantic coverage criterion used by the diagnostic. It tests whether the source-supported information in $u$ remains recoverable from $m$ rather than requiring verbatim overlap. For $|\mathcal S_i|>0$, the record-level coverage is
\begin{equation}
\operatorname{Cov}_i
=
\frac{1}{|\mathcal S_i|}
\sum_{u\in\mathcal S_i}c(u,s_i),
\label{eq:appendix-record-coverage}
\end{equation}
and the text-stage score at evolution round $r$ is
\begin{equation}
g_r^{\mathrm{text}}
=
\frac{
\sum_i\sum_{u\in\mathcal S_i}c(u,s_i)
}{
\sum_i|\mathcal S_i|
},
\qquad
\sum_i|\mathcal S_i|>0.
\label{eq:appendix-text-score}
\end{equation}
The global score measures source preservation, whereas the record-level values localize where information is lost.

For an uncovered sentence, the diagnostic extracts only atomic facts supported by that source sentence,
\begin{equation}
\mathcal A(u)=\operatorname{Decompose}(u),
\qquad
u\in\mathcal S_i,
\quad c(u,s_i)=0.
\label{eq:appendix-atomic-omissions}
\end{equation}
The records with the largest omission rates, their source-supported missing facts, and their construction trajectories form the coverage evidence $D_{\mathrm{cov}}$. This evidence explains what was lost and where the loss arose; it does not replace $r_{\Delta^\ell}$ or participate in construction-time calibration.

\paragraph{Redundancy.}
Let
\begin{equation}
\mathcal V=\bigcup_i\operatorname{Sent}(s_i)
\label{eq:appendix-retained-sentences}
\end{equation}
be the retained textual sentences for a video. Semantic similarity is used only to propose potentially overlapping pairs in $\mathcal V$. The diagnostic then compares their temporal locations and source-caption provenance before classifying them as equivalent, subsumed, or partially overlapping evidence. Semantically similar descriptions of distinct recurring events are therefore kept separate. Confirmed groups are ranked by the token capacity that consolidation would reclaim, and the highest-impact groups and their construction trajectories form the redundancy evidence $D_{\mathrm{red}}$.

\paragraph{Diagnosis and targeted revision.}
At round $r$, a bounded subset of $D_{\mathrm{cov}}$, $D_{\mathrm{red}}$, and the associated trajectories forms the phase diagnostic $D_r$. The coding agent attributes the dominant information loss or capacity waste to exactly one operator in
\begin{equation}
\{\op{Merge},\op{Allocate},\op{Compress},\op{Verify},\op{Repair}\}.
\label{eq:appendix-text-operator-set}
\end{equation}
The targeted edit modifies only that operator. All other operators, interfaces, the memory representation, and the text budget $B_s$ remain fixed. The revised program is then evaluated by $g_r^{\mathrm{text}}$, after which the guarded mutation rule in \cref{app:guarded-mutation} determines the next candidate.

\subsubsection{Image-Writer Diagnostics}

After text construction is selected, the records $\{s_i\}$ remain frozen throughout image-Writer evolution. For each retained pair $(t,I)\in\mathcal I_i$, a fixed multimodal evaluator compares image $I$ with $s_i$ and assigns
\begin{equation}
y(I,s_i)
\in
\{\textsc{Complementary},
  \textsc{Text-Overlapping},
  \textsc{Uninformative}\}.
\label{eq:appendix-image-labels}
\end{equation}
A complementary image preserves recognizable evidence not adequately recoverable from $s_i$; a text-overlapping image mainly repeats the textual record; and an uninformative image provides no sufficiently clear or useful evidence.

Visual redundancy is diagnosed with the same frozen DINO features used by $\rho_{\mathrm{DINO}}$. For each selected image, the diagnostic examines its maximum similarity to the other images in $\mathcal I_i$ and treats high-similarity pairs as candidate duplicates. The multimodal evaluator then checks whether those pairs preserve distinct details or unnecessarily repeat the same visual evidence.

Let
\begin{equation}
K=\sum_i|\mathcal I_i|
\label{eq:appendix-selected-image-count}
\end{equation}
be the total number of retained images. When $K>0$, the diagnostic reports normalized rates
\begin{equation}
R_y=\frac{N_y}{K},
\qquad
y\in
\{\textsc{Text-Overlapping},
  \textsc{Uninformative},
  \textsc{Redundant}\},
\label{eq:appendix-image-failure-rates}
\end{equation}
where $N_y$ is the number of selected image slots assigned to failure type $y$. When $K=0$, these rates are not evaluated; instead, the diagnostic records the zero-selection outcome and the corresponding retention and allocation traces. This avoids treating an empty $\mathcal I_i$ as an error when text is sufficient or no image passes retention.

The aggregate labels, redundancy evidence, affected observations, and selection traces form the image-stage diagnostic $D_r$. The image-stage score $g_r^{\mathrm{img}}$ is oriented so that higher values indicate stronger visual complementarity. A targeted revision changes exactly one of the components identified in the main text: residual-value scoring, DINO-based redundancy penalization, or the retention rule. The textual records and all non-targeted selector components remain fixed.

\subsection{Reader-Evolution Diagnostics}
\label{app:reader-evolution-diagnostics}

Reader evolution uses incorrectly answered labeled examples to diagnose failures in evidence access and organization. The memory $M$, its textual records $\{s_i\}$ and image sets $\{\mathcal I_i\}$, and the answer model $A^\star$ remain frozen.

\paragraph{Execution traces.}
For each incorrectly answered pair $(q,\hat a)$ with reference temporal interval $\hat\Delta$, let $\Xi(q)$ denote the ordered execution log of the reading workflow in the main text. It records:
\begin{enumerate}[leftmargin=*,itemsep=1pt,topsep=2pt]
    \item the input query, candidates, rankings, and textual evidence $\mathcal E_s$ produced by $\op{Retrieve}_s$;
    \item the inputs $(q,\mathcal E_s)$, the unresolved information identified by $\op{Reform}$, and its reformulated query $q'$;
    \item the image candidates, retrieval decisions, and visual evidence $\mathcal E_v$ produced by $\op{Retrieve}_v$, or the decision to skip visual retrieval and set $\mathcal E_v=\varnothing$;
    \item the inputs $(q,\mathcal E_s,\mathcal E_v)$ and the organized context $E$ produced by $\op{Reorg}$; and
    \item the final prediction $a=A^\star(q,E)$.
\end{enumerate}
The diagnostic judge receives $\Xi(q)$ together with $q$, $a$, $\hat a$, $\hat\Delta$, and the corresponding evidence available in frozen memory.

\paragraph{Failure attribution.}
The trajectory is assigned one of the three failure types defined in the main text:
\begin{enumerate}[leftmargin=*,itemsep=1pt,topsep=2pt]
    \item \emph{Text-localization failure}: relevant text associated with $\hat\Delta$ exists in $M$ but is absent from $\mathcal E_s$. The responsible operator is $\op{Retrieve}_s$.
    \item \emph{Visual-localization failure}: an image supporting $\hat a$ exists in $M$ but is absent from $\mathcal E_v$. The responsible operator is $\op{Reform}$ or $\op{Retrieve}_v$, as determined from the trajectory.
    \item \emph{Evidence-utilization failure}: relevant evidence was retrieved, but redundant, irrelevant, or poorly ordered context obscured it in $E$. The responsible operator is $\op{Reorg}$.
\end{enumerate}
If the evidence needed for $\hat a$ was never retained in $M$, the example is not labeled as a Reader localization failure. After attribution, the judge identifies the specific operator behavior that caused the error. Recurring operator-specific diagnoses and a bounded set of supporting trajectories are aggregated into the round diagnostic $D_r$. A targeted Reader edit changes only the implicated operator, while the remaining Reader operators, $M$, and $A^\star$ stay fixed. Reader candidates are evaluated by a higher-is-better QA score $g_r^{\mathrm{reader}}$.

\subsection{Guarded Mutation Algorithm}
\label{app:guarded-mutation}

Text-Writer, image-Writer, and Reader evolution use the same guarded program-mutation strategy with phase-specific diagnostics and scores. Let $P_r$ be the workflow program evaluated at round $r$, $D_r$ its bounded diagnostic feedback, and $g_r$ its evaluation score. All scores are oriented so that higher is better: text-Writer evolution measures source preservation, image-Writer evolution measures visual complementarity, and Reader evolution measures downstream QA performance.

Given the current program and diagnostic context, the coding agent proposes a targeted edit
\begin{equation}
\Delta_r=C(P_r,D_r),
\label{eq:targeted-code-edit}
\end{equation}
where the permitted scope of $\Delta_r$ is the component implicated by $D_r$. All non-targeted components, program interfaces, frozen artifacts, the memory representation, and the budget $B=(B_s,B_v)$ remain unchanged.

\subsubsection{Best-So-Far Tracking}

After evaluating the initial program, the best-so-far record is initialized as
\begin{equation}
(P_0^\star,g_0^\star)=(P_0,g_0).
\label{eq:best-program-initialization}
\end{equation}
For each subsequent round $r\geq1$, it is updated by
\begin{equation}
(P_r^\star,g_r^\star)
=
\begin{cases}
(P_r,g_r),
& g_r>g_{r-1}^\star,\\
(P_{r-1}^\star,g_{r-1}^\star),
& \text{otherwise}.
\end{cases}
\label{eq:best-program-update}
\end{equation}
Thus, the record changes only when the newly evaluated program achieves a strictly higher score.

\subsubsection{Mutation Rule}

The next candidate program is selected according to
\begin{equation}
P_{r+1}
=
\begin{cases}
P_r^\star,
& r\geq1
\ \text{and}\
 g_{r-1}-g_r>\tau_{\mathrm{rev}},\\[1mm]
\operatorname{Apply}(P_r,\eta_{\mathrm{exp}}),
& r\geq2,
\ |g_r-g_{r-1}|<\epsilon,
\ \text{and}\
 |g_{r-1}-g_{r-2}|<\epsilon,\\[1mm]
\operatorname{Apply}(P_r,\Delta_r),
& \text{otherwise},
\end{cases}
\label{eq:shared-mutation-strategy}
\end{equation}
where $\tau_{\mathrm{rev}}\geq0$ is the rollback threshold, $\epsilon>0$ is the stagnation tolerance, and $\operatorname{Apply}(P,\delta)$ applies edit $\delta$ to program $P$. The rollback branch has priority whenever its condition and the stagnation condition both hold.

\paragraph{Rollback.}
If the score decreases by more than $\tau_{\mathrm{rev}}$ relative to the preceding round, the next program is restored to the best-so-far program $P_r^\star$. This prevents a substantially degraded candidate from becoming the basis of subsequent mutations.

\paragraph{Exploration after stagnation.}
If two consecutive score changes are smaller than $\epsilon$, the coding agent proposes an exploratory edit $\eta_{\mathrm{exp}}$. Unlike $\Delta_r$, this edit need not target the component implicated by the latest diagnosis. It must nevertheless stay within the legal mutation space of the current evolution phase and preserve the phase's frozen artifacts, program interfaces, memory representation, and budget $B=(B_s,B_v)$. This permits movement beyond the latest diagnosis without coupling the text Writer, image Writer, and Reader evolution phases.

\paragraph{Targeted mutation.}
When neither rollback nor exploration is triggered, the coding agent applies $\Delta_r$ to the current program. At $r=0$, neither historical guard is available, so the targeted edit is applied. At $r=1$, only rollback can be evaluated. Exploration becomes eligible from $r=2$, when two consecutive score changes are available.

\subsubsection{Stopping and Program Selection}

Each evolution phase terminates when its phase-specific convergence criterion is satisfied or its prescribed maximum number of rounds is reached. This convergence criterion is distinct from the stagnation condition in \cref{eq:shared-mutation-strategy}: two stagnant score changes trigger exploration rather than immediate termination. If the phase terminates after round $R$, it returns
\begin{equation}
P_{\mathrm{return}}=P_R^\star
\label{eq:return-best-program}
\end{equation}
rather than the final evaluated candidate $P_R$. Consequently, every phase returns its highest-scoring observed program.

\subsection{Multi-Scale Source-Corpus Construction}
\label{app:multiscale-source-corpus}

We instantiate the multi-scale source corpus in the main text with $L=3$ temporal scales, ordered from coarsest to finest. Specifically, the full video is uniformly partitioned into
\begin{equation}
|\mathcal T^1|=16,
\qquad
|\mathcal T^2|=32,
\qquad
|\mathcal T^3|=64
\label{eq:multiscale-partitions}
\end{equation}
non-overlapping intervals. The shared video boundaries make the hierarchy nested: every interval $\Delta^\ell\in\mathcal T^\ell$ at a coarser scale is the union of two adjacent intervals at the next finer scale. This dyadic structure supports the bottom-up consolidation in \cref{eq:compact-operators}, from the 64-interval scale to the 32-interval scale and finally to the 16-interval scale.

For each interval $\Delta^\ell$, we uniformly sample 50 frames over its temporal extent and provide them to the captioner, which independently generates the source caption $c_{\Delta^\ell}$. These captions constitute the multi-scale source corpus used by the Writer; no information is propagated across intervals during caption generation, and cross-scale consolidation begins only in the subsequent writing workflow.

\begin{figure}
    \centering
    \includegraphics[width=1.0\linewidth]{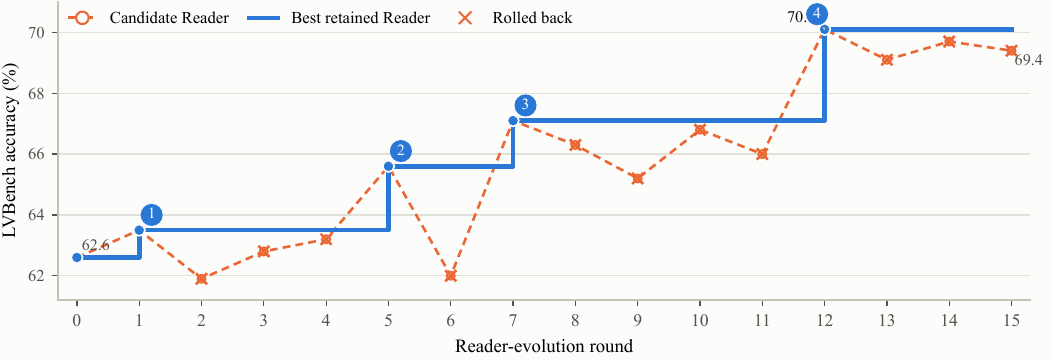}
    \caption{\textbf{Trajectory of Reader evolution.}}
    \label{fig:evo_read}
\end{figure}

\begin{figure}
    \centering
    \includegraphics[width=0.9\linewidth]{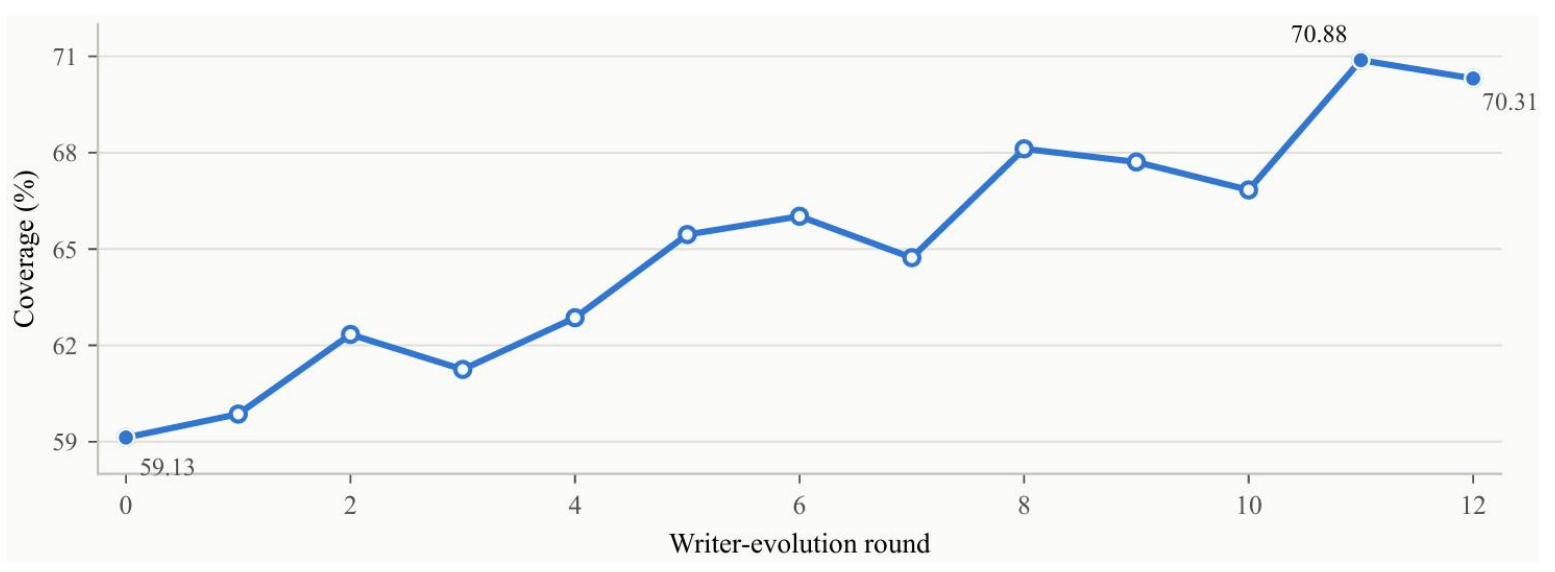}
    \caption{\textbf{Trajectory of Writer evolution.}}
    \label{fig:evo_write}
\end{figure}

\section{Evolution Trajectory}
We evolve the Writer and Reader 12 rounds and 15 rounds, respectively. The evolution trajectories are shown in Fig.~\ref{fig:evo_read} and Fig.~\ref{fig:evo_write}. To ensure stable selection of the final Reader, we evaluate the Reader obtained at each evolution round on all 30 development videos available during evolution and compute its aggregate QA accuracy over this full set. We select the checkpoint with the highest accuracy as the final evolved Reader.

\section{Storage reduction}
A key advantage of our compact video memory is that it avoids repeatedly storing and processing the complete video stream. Each video is converted into a bounded persistent memory containing: (1) a textual memory with a 20K-token budget, and (2) 160 sparsely selected visual observations. The visual observations are stored as JPEG images with the resized to $448\times448$ pixels. This representation retains temporally distributed textual and visual evidence while substantially reducing the persistent storage footprint. Once constructed, the same memory can be reused for all questions associated with the video.
  
We measure storage using the actual serialized payload rather than estimating it from the number of observations. For a set of videos $\mathcal{V}$, the original-video storage is
\begin{equation}
  S_{\mathrm{video}}
  =
  \sum_{v \in \mathcal{V}}
  \operatorname{bytes}(V_v),
\end{equation}
where $\operatorname{bytes}(V_v)$ is the on-disk size of the original compressed MP4 file. The compact-memory storage is
\begin{equation}
  S_{\mathrm{memory}}
  =
  \sum_{v \in \mathcal{V}}
  \left[
  \operatorname{bytes}\!\left(M^{\mathrm{text}}_v\right)
  +
  \sum_{j=1}^{160}
  \operatorname{bytes}\!\left(I_{v,j}^{448}\right)
  \right],
\end{equation}
where $M^{\mathrm{text}}_v$ is the final merged text serialized in UTF-8 and $I_{v,j}^{448}$ is the encoded $448\times448$ JPEG image. We exclude model parameters and temporary construction caches, since they are shared infrastructure rather than per-video persistent memory.
Based on this accounting, the compact representation occupies only $0.142\%$ of the original-video storage, yielding a $99.858\%$ reduction in persistent storage.
Thus, the proposed representation preserves sparse cross-modal evidence while remaining substantially cheaper to store, load, and reuse than the original videos.

\end{document}